\documentclass[a4paper, twocolumn, 11pt]{article}
\usepackage{geometry}
\usepackage{multirow}
\usepackage{pgfplots}
\pgfplotsset{compat=newest}
\usepackage{graphicx}
\usepackage{tabularx}
\usepackage{booktabs}
\usepackage{footmisc}
\usepackage{paralist}
\usepackage{hyphenat}
\usepackage{authblk}
\usepackage{hyperref}
\usepackage{rotating}

\providecommand{\keywords}[1]
{
  \small	
  \textbf{\textit{Keywords---}} #1
}

\newcolumntype{s}{>{\hsize=.40\hsize}X}

\newcommand{\promptore}{\textbf{PromptORE}}
\newcommand{\bcube}{B\textsuperscript{3}}
\newcommand{\mask}{\texttt{[MASK]}}
\newcommand{\sep}{\texttt{[SEP]}}
\newcommand{\cls}{\texttt{[CLS]}}

\newcommand{\bert}{\mathrm{BERT}}
\renewcommand{\S}{\mathbf{S}}
\newcommand{\dataset}{\mathcal{D}}
\newcommand{\prompt}{\mathcal{P}}
\newcommand{\labels}{\mathcal{L}}
\newcommand{\eone}{\mathbf{e1}}
\newcommand{\etwo}{\mathbf{e2}}

\title{PromptORE - A Novel Approach Towards Fully Unsupervised Relation Extraction}

\author[1, 2]{Pierre\hyp{}Yves Genest}
\author[2]{Pierre\hyp{}Edouard Portier}
\author[2]{El\H{o}d Egyed\hyp{}Zsigmond}
\author[1]{Laurent\hyp{}Walter Goix}
\affil[1]{Alteca, 88 Boulevard des Belges, 69006 Lyon, France} 
\affil[2]{Univ Lyon, INSA Lyon, LIRIS, CNRS UMR5205, 20 Avenue Einstein, 69621 Villeurbanne, France}
\affil[ ]{\textit {\{pygenest,lwgoix\}@alteca.fr} \textit {\{pierre-edouard.portier,elod.egyed-zsigmond\}@insa-lyon.fr}}

\date{October 17, 2022}

\begin{document}

\maketitle
\begin{abstract}
Unsupervised Relation Extraction (RE) aims to identify relations between entities in text, without having access to labeled data during training. This setting is particularly relevant for domain specific RE where no annotated dataset is available and for open-domain RE where the types of relations are \textit{a priori} unknown. Although recent approaches achieve promising results, they heavily depend on hyperparameters whose tuning would most often require labeled data. To mitigate the reliance on hyperparameters, we propose PromptORE, a "Prompt-based Open Relation Extraction" model. We adapt the novel prompt-tuning paradigm to work in an unsupervised setting, and use it to embed sentences expressing a relation. We then cluster these embeddings to discover candidate relations, and we experiment different strategies to automatically estimate an adequate number of clusters. To the best of our knowledge, PromptORE is the first unsupervised RE model that does not need hyperparameter tuning. Results on three general and specific domain datasets show that PromptORE consistently outperforms state-of-the-art models with a relative gain of more than $40\%$ in \bcube{}, V-measure and ARI. Qualitative analysis also indicates PromptORE’s ability to identify semantically coherent clusters that are very close to true relations \footnote{Source code is available at \url{https://github.com/alteca/PromptORE}.}.
\end{abstract}


\section*{Disclaimer}
This paper was published as a part of the Proceedings of the 31st ACM International Conference on Information and Knowledge Management (CIKM '22), October 17--21, 2022, Atlanta, GA, USA.
This document is the accepted/author version. The published version is accessible at \url{https://doi.org/10.1145/3511808.3557422}.

\section{Introduction}
Information Extraction models aim to extract the meaningful information from text, that is, entities and relations between these entities. The resulting network of relations can then be transformed into knowledge graphs that are used in multiple downstream tasks such as recommender systems \cite{Guo2020}, logical reasoning \cite{Chen2020} or question answering \cite{Huang2019}. Information Extraction is usually seen as a two-step process: 
\begin{inparaenum}
    \item Named Entity Recognition and
    \item Relation Extraction.
\end{inparaenum}
In this paper, we focus on Relation Extraction, which consists in identifying the relation between two entities in the context of a piece of text.

Relation Extraction (RE) is often seen as a supervised task \cite{Ji2021}, thus relying on datasets labeled with a predefined set of relations. However, this setting can be restrictive for some applications, especially domain-specific RE lacking annotated data or open-domain RE where we do not know in advance the relations expressed in the dataset. Therefore, more flexible paradigms have been proposed such as distant-supervision \cite{Mintz2009, Riedel2010}, which tries to automatically annotate data; few-shot learning \cite{Perez2021}, which learns from a very small set of labeled instances; or unsupervised learning. In particular, unsupervised RE (also called OpenRE) does not require a training dataset with labeled relations and assume no prior knowledge about expected types of relations.

Several recent OpenRE approaches obtain interesting results on datasets containing tens or hundreds of relation types \cite{Hu2020, Wu2019, Lou2021}. They often try to compute a vector representation of the relation expressed in the sentence (also called \textit{relation embedding}) and then cluster all the embeddings to identify groups of similar relations. Most of these methods rely on hyperparameters (e.g. number of epochs, regularization, early stopping, number of relations types, ...) that have a significant impact on their overall performance. However, tuning these hyperparameters most often requires access to labeled data, thus limiting the applicability of such models in a real-world unsupervised scenario.

\smallskip
We therefore propose \promptore{}, a "Prompt-based Open Relation Extraction" model, which relies on one hyperparameter at most: the target number $k$ of relation types to be extracted. Our experiments show that even when no educated guess can be made about $k$, an efficient estimate can easily be obtained in an automatic way. Thus, to the best of our knowledge, \promptore{} is the first proposal for an unsupervised Relation Extraction system that can operate in a fully unsupervised setting.

To achieve this, we first compute, for each instance (a piece of text) of a dataset, a relation embedding that represents the relation expressed in the instance. Contrary to previous approaches that fine-tuned BERT \cite{Hu2020, Wu2019, Zhao2021}, we use the novel prompt-tuning paradigm. Prompt-tuning replaces the usual training by designing a prompt (i.e., a text that is inputted to BERT), able to elicit as much information as possible from the Pretrained Language Model. Prompt-tuning is already used in few-shot RE \cite{Lv2022, Chen2022, Gong2021a, Tan2022}. We propose to go further and adapt this paradigm to work in a fully unsupervised way. Prompt-tuning has many benefits:
\begin{inparaenum}
\item it does not involve training or fine-tuning BERT, thus removing a significant number of hyperparameters,
\item the proposed encoder is extremely simple, yet
\item we show that these prompt-based relation embeddings provide better results than current state-of-the-art methods.
\end{inparaenum}
Usual clustering algorithms are then applied to group together the embeddings in order to discover relation types.

\smallskip
Let us summarize our main contributions:
\begin{itemize}
    \item We propose \promptore{}, a novel OpenRE model that minimizes the number of hyperparameters and provides clear ways to tune its only hyperparameter $k$ in a strict unsupervised setting.
    \item We adapt the prompt-tuning paradigm to an unsupervised setting, which allows us to leverage more expressive embeddings than previous entity-pair representations \cite{Hu2020, Wu2019, Luan2019}.
    \item We show that this model outperforms consistently previous state-of-the-art approaches on three different datasets, covering both general and specific domains. We also demonstrate that the predicted clusters are semantically coherent and very close to the true relations.
\end{itemize}

\section{Related Work}\label{sec:related_work}
For the sake of clarity, we use \textit{relation} and \textit{relation type} interchangeably, and we define them as a concept linking two entities, for example \texttt{married\_to}, \texttt{born\_in}, \texttt{located\_in}; but we distinguish \textit{relation instance}, which represents the realization of a relation, that is, a piece of text expressing the relation.

Relation Extraction aims to discover the binary relation that links two entities mentioned in a text. RE allows to extract triples of the form ($\eone$, relation, $\etwo$) that can be used thereafter to build for instance a knowledge graph. Even though Relation Extraction from documents is the most general paradigm, the majority of models focus on extraction within a single sentence ignoring inter-sentence relations \cite{Han2019, Han2020}. Recent approaches follow a two-step process \cite{Han2020, Lin2020, Luan2019, Soares2019}:
\begin{enumerate}
    \item Relation Embedding, which computes a vector representation of the relation instance,
    \item Relation Classification.
\end{enumerate}
To compute relation embeddings, word-embedding models are often used, such as GLoVe \cite{Pennington2014}, ELMO \cite{Peters2018}, Bi-LSTM embeddings \cite{Luan2019}, or more recently BERT embeddings \cite{Vaswani2017, Devlin2019}. In the general case, relation classification is seen as a supervised task therefore needing labeled datasets, where entities have been extracted and labels describing the relation are available.

\paragraph{Supervised Relation Extraction}
Recent models implement a joint entity and relation extraction scheme \cite{Ji2021}. Luan et al. \cite{Luan2019} propose a multi-task learning approach that simultaneously optimizes a BERT-based entity extractor, relation extractor and coreference resolution model. Wadden et al. \cite{Wadden2019} add event triggers and roles detection into the same multi-task framework. Lin et al. \cite{Lin2020} improve these models by incorporating external constraints on entity types and relations. Zhong et al. \cite{Zhong2021} show nevertheless that a simple pipelined approach outperforms complex multi-task models, thanks to a markup-based encoding of the sentence. However, these supervised approaches rely on large labeled datasets, available only for generic language corpora and a few specialized domains. Therefore, other works aim at reducing this reliance on such annotated data.

\paragraph{Distantly-supervised Relation Extraction}
Distant-supervision \cite{Mintz2009, Riedel2010} tackles this problem by automatically annotating texts based on external knowledge bases. This annotation process creates large scale datasets that are characterized by a high level of noise. Zhang et al. \cite{Zhang2021c} identify two types of noise: intra-dictionary bias (spurious entity/relation annotations due to wrong entity linking) and inter-dictionary bias (non exhaustiveness of knowledge bases, meaning that some entities/relations cannot be labeled). Most works focus on trying to reduce and mitigate these biases \cite{Zhang2019, Zhang2021a, Zhang2021c, Zheng2019}. However, distant-supervision does not apply to very specific domains that lack large knowledge bases.

\paragraph{Few-Shot Relation Extraction}
This approach aims to learn a relation extraction model from the least amount of labeled data. Models focus on relation classification: they suppose the existence of a relation between two entities \cite{Perez2021}, ignoring the case of sentences mentioning unrelated entities. Snell et al. \cite{Snell2017} propose to use prototypical networks to determine a prototype for each relation, and predict the relation by measuring the distance between the relation instance embedding and each prototype. Ren et al. \cite{Ren2020} and Zhao et al. \cite{Zhao2021} propose to improve this method using transfer-learning with general domain labeled datasets.

Very recent few-shot methods consider the use of prompt-tuning with BERT and more broadly Pretrained Language Models (PLMs) \cite{Gao2021}, as it allows for more efficient learning in low-resource setting \cite{Liu2021a}. Prompt-tuning replaces fine-tuning by designing a "prompt", that is, a piece of text containing the special \mask{} token, and ask a PLM such as BERT to predict the embedding of this \mask{} token. This embedding is then compared with a set of target tokens (that can be seen as relation prototypes) to determine the relation expressed in this sentence \cite{Lv2022, Chen2022, Gong2021a, Tan2022}. Efforts are focused in optimizing the prompt $\prompt$ and selecting a set of target tokens effective at representing the relations. In particular, Jiang et al. \cite{Jiang2020} propose text-mining and paraphrasing-based methods to generate prompts.

The current major limitation with few-shot RE is the closed-world hypothesis, which stipulates that relations must be known in advance (although recent papers start to explore \textit{none-of-the-above} prediction \cite{Gao2019, Lv2022}).

\paragraph{Unsupervised Relation Extraction}
Unsupervised RE (or OpenRE \cite{Banko2007}) aims to extract relations without having access to a labeled dataset during training. Methods can be divided in two subgroups: (1) triples extraction and (2) relation typing. Banko et al. \cite{Banko2007} extract triple candidates using syntactic rules and refine the candidates with a trained scorer. Saha et al. \cite{Saha2018} propose to simplify conjunctive sentences to improve triples extraction. More recently, neural networks and word-embedding were applied to solve this task \cite{Stanovsky2018, Cui2018}, requiring a general domain annotated dataset to pretrain their model. Finally, Roy et al. \cite{Roy2019} propose an ensemble method to aggregate results of multiple OpenRE models. These triples extraction approaches rely on surface forms, which makes it hard for them to group instances that express the same relation using very different words and syntax.

To solve this problem, Yao et al. \cite{Yao2011} propose instead to learn a relation classifier, using Latent-Dirichlet Allocation \cite{Blei2003}, a generative probabilistic model. The majority of these relation typing methods rely on relation embeddings: first, they compute an embedding, which encodes the underlying relation, second they use this embedding to identify groups of relation instances. The earliest methods use syntactic and semantic features \cite{Yao2011, Yao2012, Marcheggiani2016}. Elsahar et al. \cite{Elsahar2017} add word-embedding features based on GloVe \cite{Pennington2014}, apply dimensionality reduction methods, and an agglomerative clustering model to identify clusters of relation instances. Marcheggiani et al. \cite{Marcheggiani2016} use a fill-in-the-blank task: they mask one entity and try to predict it using a Variational Auto-Encoder (VAE) \cite{Kingma2014}, proving the benefit of generating a supervision signal. This method is further improved by adding two regularization losses to limit overfitting \cite{Simon2019} and by finding a more effective formulation of the VAE task \cite{Yuan2021}. Tran et al. \cite{Tran2020b} however succeed to outperform VAE approaches only using entity types as their relational embeddings.

Hu et al. \cite{Hu2020} adopt an other supervision signal: they compute pseudo labels using a k-means clustering on relation embeddings, and train a classifier to reproduce these pseudo labels, allowing them to fine-tune a BERT model. As an alternative, Wu et al. \cite{Wu2019} propose to learn a distance metric representative of the relations (using Siamese neural networks \cite{Chopra2005}), to compare pairs of instances. This metric is learned on an annotated dataset, and applied to unlabeled data to identify instances expressing similar relations. Lou et al. \cite{Lou2021} use ranked list loss \cite{Wang2019a} as an alternative to Siamese neural networks. Finally, a tendency of recent unsupervised RE methods is to use transfer-learning: learning some relation embeddings or metrics on general domain annotated datasets and try to adapt them to unsupervised data \cite{Zhao2021, Lou2021, Wu2019}. Compared to triples extraction, relation typing assumes that there is always a relation between the two entities, which can be seen as a limitation.

To allow evaluation of such OpenRE models, previous works tend to train them on labeled datasets and compare their predictions with ground truth relations using external clustering evaluation metrics such as V-measure \cite{Rosenberg2007}, Adjusted Rand Index \cite{Hubert1985, Steinley2004} or \bcube{} \cite{Baldwin1998}.

\paragraph{Are current OpenRE models truly unsupervised ?} Although Open-RE models extract relations from unannotated datasets, we argue that they are not truly unsupervised approaches: the main problem is hyperparameter tuning. All these approaches rely extensively on hyperparameters that need to be adjusted: number of epochs/iterations \cite{Marcheggiani2016, Yuan2021, Tran2020b, Hu2020, Wu2019, Lou2021}, learning rate, regularization \cite{Simon2019}, entity types \cite{Tran2020b}, early-stopping \cite{Tran2020b}, etc., and most importantly the number of relations $k$ the model is supposed to extract \cite{Marcheggiani2016, Yuan2021, Tran2020b, Hu2020, Wu2019, Lou2021, Simon2019, Elsahar2017}. In a real unsupervised setting these hyperparameters are extremely hard to determine, and cited papers do not present satisfactory methods to estimate them without labeled data. Therefore, we conclude that these mentioned approaches are not fully unsupervised when it comes to hyperparameter tuning, which in our opinion, restricts their use in a real-world application.

As a result, it motivates us to define more precisely the unsupervised RE setting as \textit{learning a RE model and tuning its hyperparameters using only unlabeled data}.

\section{Proposed Model}
\promptore{} aims to extract the binary relation $r$ between two already known entities $\eone$ and $\etwo$ present in the same sentence\footnote{As previous works, we focus on sentence RE, even though we are aware that some relations may be missed.}. More precisely, as we follow an unsupervised setting, the first objective of \promptore{} is to group instances expressing the same relation $r$, without having access to labeled data during training and hyperparameter tuning. Our second objective is to minimize the number of hyperparameters needed by \promptore{} and to provide clear procedures to adjust them without annotated data.

\begin{figure}[tb]
    \centering
    \includegraphics[width=0.45\textwidth]{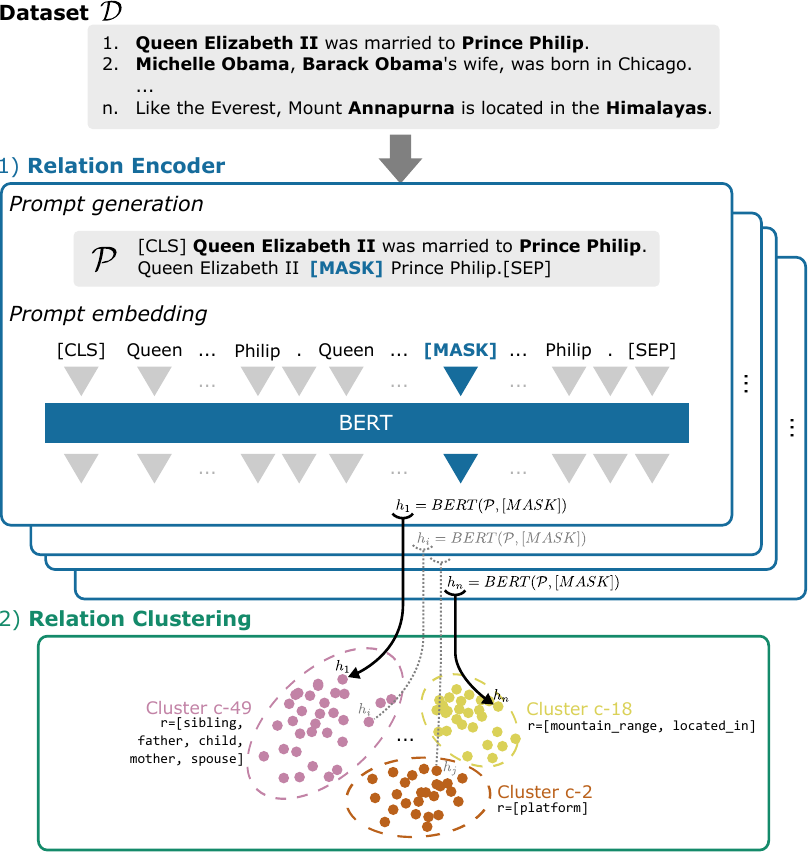}
    \caption{Overview of PromptORE.}
    \label{fig:overview}
\end{figure}

To achieve these goals, we suppose we have access to a dataset $\dataset$ (see Figure \ref{fig:overview}) containing instances with the following properties:
\begin{itemize}
    \item An instance is described with a triple $(\S, \eone, \etwo)$, where $\S=[t_0,...,t_{s-1}]$ is the instance text composed of tokens\footnote{\textit{Token} as defined by BERT \cite{Devlin2019}: punctuation, word or part of word.}, $\eone=[t_{start(\eone)},..., t_{end(\eone)}]$ and $\etwo=[t_{start(\etwo)},..., t_{end(\etwo)}]$ are two entities identified by their indexes in $\S$.
    \item We suppose that $\eone$ and $\etwo$ have already been extracted (but not typed).
    \item In the instance text $\S$, $\eone$ and $\etwo$ are linked by a binary relation $r$. As previous approaches, we do not consider the case where there is no relation between $\eone$ and $\etwo$.
    \item We do not have access to any relation label during training and hyperparameter tuning.
\end{itemize}
$\mathcal{R}$ is the set of the $k$ relations contained in $\dataset$. We consider that we have no information about the relations in $\mathcal{R}$ (e.g. their labels, their linked entity types, etc.). Regarding $k$, it can either be given by the user or automatically estimated by methods described in section \ref{sec:relation_clustering}.

\medskip
As shown in Figure \ref{fig:overview}, \promptore{} is composed of two main modules (similarly to \cite{Hu2020, Zhao2021}):
\begin{enumerate}
    \item \textit{Relation Encoder}. This module computes a vector representation of the relation that is expressed in the current instance. To do that, we apply a modified prompt-tuning method to leverage BERT embeddings.
    \item \textit{Relation Clustering}. It clusters the relation embeddings of the whole dataset, in order to identify groups of instances that are expected to express the same type of relation\footnote{In practice, clusters may not be perfectly pure: they may contain instances expressing different relations, e.g. clusters c-18 or c-49 in Figure \ref{fig:overview} or Table \ref{tab:cluster_content}.}.
\end{enumerate}

\subsection{Relation Encoder}
This module aims to compute a vector representation (or \textit{relation embedding}) of the relation expressed between $\eone$ and $\etwo$ in the current sentence $\S$. We want this relation embedding to be representative of the underlying relation: if the relation embeddings of two instances are close (relative to a certain distance metric), these instances convey, most probably, the same relation. In other words, the Relation Encoder aims to abstract the notion of relation instance, to provide embeddings that are easier to compare.

In recent papers \cite{Hu2020, Zhao2021, Liu2021, Lou2021, Yuan2021, Zhong2021, Wadden2019}, relation embeddings are computed using Pretrained Language Models such as BERT \cite{Devlin2019, Vaswani2017} or RoBERTa \cite{Liu2019}. BERT (and RoBERTa) takes as input some tokenized text, and computes for each input token an embedding, which is representative of the token itself and its context of use. BERT also contains a Masked Language-Model (MLM) head, which allows it to predict the most probable tokens associated with an embedding. In addition to "real word" tokens, BERT uses some special tokens:
\begin{itemize}
    \item \cls{} and \sep{}. By convention, \cls{} needs to be inserted at the start of the text, and \sep{} indicates the end of the text.
    \item \mask{}. It represents a token that is hidden/unknown, and BERT will try to compute a satisfactory embedding. Then, using the MLM head, BERT can predict the most probable tokens. Thanks to this \mask{} token, BERT can auto-complete sentences or generate text.
\end{itemize}
We define $h=\bert(\S, \mask)$ as the BERT-embedding of the token \mask{}, in the context of $\S$ (a piece of text that must contain exactly one \mask{} token). Finally, for the sake of simplicity, variables/parameters (written with a bold name), are automatically replaced by their value in a quoted string. For example, if $\mathbf{a}=\textit{One}$, \texttt{"$\mathbf{a}$, Two"} corresponds to \textit{One, Two}.

Previous works \cite{Hu2020, Zhao2021, Liu2021, Lou2021, Yuan2021, Zhong2021, Wadden2019} use an entity-pair representation paradigm to compute relation embeddings. We however opt for another technique: prompt-based encoding.

\paragraph{Prompt-Tuning}
The idea behind prompt-tuning is to benefit from BERT's ability to predict masked tokens (\mask{} tokens). It is already used in few-shot RE by \cite{Gong2021a, Chen2022, Lv2022, Tan2022}. It can be summarized as follows:
\begin{enumerate}
    \item Design a prompt $\prompt$, which is a sequence of tokens that includes one \mask{} token. For instance, Lv et al. \cite{Lv2022} use the template \texttt{$\prompt(\S, \eone, \etwo)=$ "\cls{} $\S$ In this sentence, $\eone{}$ is the \mask{} of $\etwo{}$.\sep{}"}.
    \item Identify a set of label tokens $\labels$ that represents each relation.
    \item \sloppy Predict the \mask{} embedding in the context of $\prompt{}$ using BERT: $h=\bert(\prompt{}, \mask{})$.
    \item With this embedding, compute the probability to predict each label token $l\in\labels$ thanks to the MLM head of BERT.
    \item Select the relation represented by the label token $l$ with the highest probability.
\end{enumerate}
Prompt-tuning does not require to fine-tune BERT, but necessitate to design an optimal prompt $\prompt$ and a set of label tokens $\labels$. They are usually adjusted using a labeled dataset and therefore cannot be applied directly to our unsupervised Relation Encoder.

\paragraph{Unsupervised Prompt-based Relation Encoder}
To adapt prompt-tuning for unsupervised RE, we propose to remove the set of label tokens $\labels$, and use the simplest prompt $\prompt$ possible. Our proposed prompt template is:
\begin{equation}
    \label{eq:prompt}
    \prompt(\S, \eone, \etwo) = \texttt{"\cls{} $\S$ $\eone$ \mask{} $\etwo$.\sep{}"}
\end{equation}
with $\S$ the instance text, $\eone$ (resp. $\etwo$) the text of the first (resp. second) entity. For example, given the sentence $\S=\,$\textit{Queen Elizabeth II was married to Prince Philip.}, and entities $\eone=\,$\textit{Queen Elizabeth II} and $\etwo=\,$\textit{Prince Philip}, the prompt is:

\begin{center}
    $\prompt{}=$ \textit{\cls{} Queen Elizabeth II was married to Prince  Philip. Queen Elizabeth II \mask{} Prince Philip.\sep{}}
\end{center}

As we remove $\labels$, we decide to use the \mask{} BERT embedding as our relation embedding. Thus, the Relation Encoder process is the following (as shown in Figure \ref{fig:overview}):
\begin{enumerate}
    \item Apply the template defined in eq. (\ref{eq:prompt}) to generate a prompt for the current instance.
    \item Predict the embedding of the \mask{} token with BERT, and use it as our relation embedding: $h=\bert(\prompt,\mask{})$.
\end{enumerate}

\paragraph{Alternative Prompts} \label{sec:alternative_prompts} If we analyze $\prompt$, we can see that BERT will likely fill the \mask{} token with a verb, as it is trained to produce grammatically correct sentences. It raises questions: can all relations be expressed with a \textit{verb}, and with a \textit{single} word? We did an analysis on the $9\,892$ Wikidata relations with surface forms:
\begin{itemize}
    \item More than 75\% of relations need 2 words or more to be expressed (e.g. acceptable surface forms for \texttt{birth\_place} are \textit{born in} or \textit{the birth place of}).
    \item Surface forms usually contain a root word (noun, verb) accompanied by tool words. $92\%$ of the root words are nouns, and only $6.7\%$ verbs. The most common tool words are: \textit{of, in, by, the, a, to}.
\end{itemize}

It can be therefore interesting to consider alternative prompts that encourage BERT to predict a noun (\cite{Lv2022} also focus on noun prediction). In addition, Lv et al. \cite{Lv2022} introduce a prefix to their prompt: \texttt{"In this sentence"}. Therefore, we define alternative prompt templates aiming at predicting noun, and with various prefixes:
\begin{enumerate}
    \item[$\prompt'_{1}(\S,\eone,\etwo) =$] \texttt{"\cls{} $\S$ $\eone$ is the \mask{} of $\etwo$.\sep{}"}
    \item[$\prompt'_{2}(\S,\eone,\etwo) =$] \texttt{"\cls{} $\S$ In this sentence, $\eone$ is the \mask{} of $\etwo$.\sep{}"}
    \item[$\prompt'_{3}(\S,\eone,\etwo) =$] \texttt{"\cls{} $\S$ We deduce that $\eone$ is the \mask{} of $\etwo$.\sep{}"}
\end{enumerate}
$\prompt'_{2}$ is the same as \cite{Lv2022}. However, we cannot choose the optimal prompt from the previous ones nor use automatic methods to generate prompts (such as \cite{Jiang2020}) as they require access to labeled data. Therefore, the main results of \promptore{} are computed using $\prompt$, the simplest prompt of all. In a second phase, we will analyze \promptore{}'s performances with these alternative prompts.

\subsection{Relation Clustering}\label{sec:relation_clustering}
We can measure the similarity between two BERT embeddings using an euclidian distance as these embeddings are normalized \cite{Devlin2019}. Similarly to previous works \cite{Elsahar2017, Hu2020, Zhao2021}, we cluster the relation embeddings computed on the entire dataset $\dataset{}$ to find groups of instances, and we expect these clusters to be good candidate relations.

\paragraph{K-Means Clustering} If we know in advance the number of relations $k$, we propose to use a simple k-means clustering \cite{Lloyd1957,Observations1967}.

\paragraph{Clustering without $k$} \label{sec:without_k} The most general case is however that we do not know $k$. To tackle this problem, we can take two points of view: use clustering models that do not require a predefined number of clusters, or estimate $k$ automatically and use it with a regular clustering method.

For the first point of view, multiple models are available, the main ones being Agglomerative Clustering (HAC) \cite{Sibson1973}, DBSCAN \cite{Ester1996}, OPTICS \cite{Ankerst1999} or Affinity Propagation \cite{Frey2007}. Nevertheless, most cannot be applied in our case: DBSCAN and HAC need other hyperpameters (such as density) and Affinity Propagation does not scale well to big datasets. Therefore, we propose to use OPTICS.

To estimate the number of clusters, we propose to implement the Elbow Rule \cite{Thorndike1953}. We select the silhouette coefficient \cite{Rousseeuw1987} as our internal metric\footnote{A metric that does not rely on external data such as labels.} to measure the quality of the clustering. Intuitively, we expect that increasing the number of clusters will improve the value of the silhouette coefficient since there are more parameters to explain the data. However when we have more clusters than the actual number of relations, the silhouette will most likely grow more slowly as we can only subdivide actual relations. The Elbow Rule tries to find the "elbow", which is the optimal trade-off between a reasonable number of clusters and a high silhouette coefficient. It can be done visually, but automatic methods are also available \cite{Kolesnichenko2021}. With this estimation, we can use any clustering algorithm, in particular k-means.

\section{Experiments}
\subsection{Datasets \& Evaluation Metrics}
To evaluate \promptore{} we exploit labeled datasets, the labels being only used during evaluation. The first dataset we choose is FewRel \cite{Han2018}\footnote{\label{note:fewrel}Data can be downloaded from: \url{https://github.com/thunlp/FewRel}.}. This dataset is composed of text taken from Wikipedia pages that has been automatically annotated by aligning the text with Wikidata triples (distant-supervision setting),  then manually checked for each instance. FewRel contains 80 relations, with 700 instances each (20 others relations are available in the test set, which is kept private). Therefore the dataset is composed of $56\,000$ instances. One important fact is that the dataset contains at most one instance for each pair of entities.

FewRel is a general domain dataset, but we also want to evaluate \promptore{} on more specific domains. Our second dataset is FewRel NYT \cite{Gao2019}\footref{note:fewrel}. This time, the text is taken from newspapers articles of the New-York Times. It is also automatically annotated using Wikidata, and manually checked. FewRel NYT contains 25 different relations, with 100 instances each.

Our third dataset is FewRel PubMed \cite{Gao2019}\footref{note:fewrel}. The text comes from PubMed, a database of biomedical literature. It is also automatically annotated (this time with the UMLS knowledge base) and manually checked. It is composed of 10 relations, with 100 instances each.

\smallskip
Traditional classification metrics such as accuracy, precision, recall or f1-score cannot be used to evaluate and compare \promptore{}'s performances as there is no direct link between our cluster ids and the true relation ids. Therefore, similarly to previous works \cite{Yao2011, Simon2019}, we use three external clustering metrics: \bcube{} \cite{Baldwin1998}, V-measure (V) \cite{Rosenberg2007} and Adjusted Rand Index (ARI) \cite{Hubert1985, Steinley2004}. They each take a different point of view: ARI is based on pairwise similarity (enumerating all pair of instances), \bcube{} on one instance versus the dataset and the V-measure on clusters.

\textit{ARI} is adjusted for chance, meaning that a random clustering will reliably lead to a score close to 0.

\textit{V-measure} defines the notions of homogeneity and completeness of clusters. A cluster is homogeneous if it contains only instances of the same relation, and a cluster is complete if it contains every instances of a relation. The V-measure corresponds to the harmonic mean of homogeneity and completeness.

Finally, \textit{\bcube{}} provides definitions for recall and precision, allowing to compute a f1-score. V-measure tends to penalize more small impurities in a pure cluster than impurities in a less pure cluster where \bcube{} has a more linear behavior \cite{Simon2019}.

\subsection{Baselines}
We compare \promptore{} with the state-of-the-art (SOTA) approach \textbf{SelfORE} \cite{Hu2020}, and two previous 
approaches based on Variational Auto-Encoders (\textbf{Etype+} \cite{Tran2020b} and \textbf{UIE-PCNN} \cite{Simon2019}).

\textbf{SelfORE} encodes instances with BERT, clusters these embeddings with an adaptive clustering method to generate pseudo labels that are finally used to train a classifier.

\textbf{UIE-PCNN} encodes instances with a Piecewise Convolutional Neural Network (PCNN) \cite{Zeng2015}, and uses a Variational Auto-Encoder (VAE) to classify instances in an unsupervised way. Additionally they propose two regularization losses (skewness and dispersion) to fight against the VAE's tendency to predict a single relation or a uniform distribution. Since \textbf{UIE-PCNN} relies on PCNN, an older embedding method, we propose to replace it with a BERT model (similarly to \cite{Tran2020b}), and we call this method \textbf{UIE-BERT}.

\textbf{Etype+} shows that using only entity types to encode instances provides better results than \textbf{UIE-PCNN}. They propose a simple typing schema for the entities: \textit{Organization, Person, Location, Miscellaneous}. We expect the performances to be lower on domain-specific datasets (FewRel NYT and PubMed).

Finally, let us recall some metrics properties to interpret the experimental results: if a model always predicts the most frequent class V-measure and ARI will be equal to 0. If a model predicts a random distribution, ARI will be close to 0 (as ARI is adjusted for chance).

\subsection{Implementation Details}
All baselines are trained with the hyperparameter values determined by their authors. For \textbf{SelfORE}, we use their publicly available implementation, and for \textbf{Etype+} and \textbf{UIE-PCNN} we use the implementation of Tran et al. \cite{Tran2020b}. The baselines are trained knowing the correct number of relations $k$ (i.e., 80 for FewRel, 25 for FewRel NYT and 10 for FewRel PubMed).

For \promptore{}, we suppose we do not know $k$, except in section \ref{sec:global_comparison}. Besides, we use the \texttt{bert-base-uncased} model to initialize BERT's weights. We also use a model with RoBERTa embeddings (using the \texttt{roberta-base} pretrained model). There are no hyperparameters to adjust.

We use the \texttt{scikit-learn} implementation of V-measure and Adjusted Rand Index, and the Hu et al. \cite{Hu2020} implementation of \bcube{}.

\section{Results}
\subsection{Comparison with Previous SOTA Models} \label{sec:global_comparison}
\begin{table*}
    \small
    \centering
    \caption{Results of PromptORE and previous SOTA models on three datasets. PromptORE knows the number of relations $k$.}
    \label{tab:global_results}
    \begin{tabular}{rc|l|ccc|ccc|c}
        \toprule
        \multicolumn{2}{r|}{\multirow{2}{*}{Dataset}} & \multirow{2}{*}{Model} & \multicolumn{3}{c|}{\bcube{}} & \multicolumn{3}{c|}{V-measure} & \multirow{2}{*}{ARI} \\
        & & & Prec. & Rec. & F1 & Hom. & Comp. & F1 & \\
        \midrule
        \multirow{6}{*}{FewRel \cite{Han2018}} & \multirow{6}{*}{$k=80$}
        & UIE-PCNN \cite{Simon2019} & 5.20 & 6.78 & 5.89 & 21.1 & 21.6 & 21.3 & 4.86 \\
        & & UIE-BERT & 1.25 & 100 & 2.47 & 0 & 100 & 0 & 0 \\
        & & EType+ \cite{Tran2020b} & 7.46 & 7.99 & 13.7 & 33.3 & 79.1 & 47.9 & 8.44 \\
        & & SelfORE \cite{Hu2020} & 24.4 & 36.3 & 29.2 & 50.4 & 56.6 & 53.2 & 24.4 \\
        & & PromptORE (RoBERTa) & 47.8 & 47.9 & 47.9 & 71.2 & 72.5 & \textbf{71.8} & \textbf{43.7} \\
        & & PromptORE (BERT) & 48.7 & 48.8 & \textbf{48.8} & 71.0 & 72.7 & \textbf{71.8} & 43.4 \\
        \midrule
        \multirow{6}{*}{FewRel NYT \cite{Gao2019}} & \multirow{6}{*}{$k=25$}
        & UIE-PCNN \cite{Simon2019} & 7.31 & 27.1 & 11.5 & 9.58 & 15.8 & 11.9 & 3.09 \\
        & & UIE-BERT & 4.00 & 100 & 7.77 & 0 & 100 & 0 & 0 \\
        & & EType+ \cite{Tran2020b} & 11.0 & 92.6 & 19.6 & 23.0 & 84.9 & 36.2 & 7.82 \\
        & & SelfORE \cite{Hu2020} & 32.4 & 48.1 & 38.7 & 50.0 & 58.9 & 54.1 & 26.8 \\
        & & PromptORE (RoBERTa) & 62.6 & 65.3 & 63.9 & 75.7 & 78.1 & 76.8 & \textbf{57.3} \\
        & & PromptORE (BERT) & 63.7 & 66.6 & \textbf{65.1} & 76.5 & 79.5 & \textbf{78.0} & 56.9 \\
        \midrule
        \multirow{6}{*}{FewRel PubMed \cite{Gao2019}} & \multirow{6}{*}{$k=10$}
        & UIE-PCNN \cite{Simon2019} & 14.4 & 45.2 & 21.9 & 10.3 & 19.2 & 13.5 & 7.23 \\
        & & UIE-BERT & 10.0 & 100 & 18.2 & 0 & 100 & 0 & 0 \\
        & & EType+ \cite{Tran2020b} & 10.0 & 100 & 18.1 & 0 & 100 & 0 & 0 \\
        & & SelfORE & 53.7 & 66.1 & 59.3 & 58.8 & 68.7 & 63.4 & 45.4 \\
        & & PromptORE (RoBERTa) & 73.7 & 73.2 & 73.5 & 76.5 & 77.2 & 76.9 & 68.1 \\
        & & PromptORE (BERT) & 77.6 & 77.2 & \textbf{77.4} & 81.0 & 81.2 & \textbf{81.1} & \textbf{73.8} \\
        \bottomrule
    \end{tabular}
\end{table*}

In this section only, to allow a fairer comparison with previous approaches, \promptore{} knows $k$, the number of different relations, and a k-means clustering is used. Table \ref{tab:global_results} shows the results of the models on our three datasets. \promptore{} consistently outperforms \textbf{SelfORE}, the previous state-of-the-art method, with approximately 19\% more in \bcube{}, 18\% in V-measure and 19\% in ARI on FewRel. It represents a relative gain in performance of more than 40\%. The performance gap is even more important with \textbf{UIE-BERT}, \textbf{UIE-PCNN} and \textbf{EType+}. We observe similar conclusions with the two other datasets.

When we look more closely, we notice that \textbf{UIE-BERT} \cite{Simon2019} obtains very poor results on all three datasets: it always predicts the same relation. Strangely, the authors of \cite{Simon2019} proposed two regularization losses to avoid precisely this situation, but this problem with \textbf{UIE-PCNN/BERT} is also observed by \cite{Tran2020b, Yuan2021}. We believe this is due to the hyperparameter that controls the balance between classification and regularization losses, which needs to be fine-tuned specifically for each dataset.

\textbf{SelfORE} has been evaluated on the FewRel dataset multiple times \cite{Zhao2021, Zhang2021b, Lou2021}, and it seems at first glance that the results obtained by these papers are better than ours (with an F1 \bcube{} between 45-55\% instead of 25-30\%). However, in these cases \textbf{SelfORE} was evaluated on a test set of FewRel with only 16 relations and $11\,200$ instances \cite{Lou2021}; or a subset of $1\,600$ instances with 16 relations \cite{Zhao2021, Zhang2021b}. Using the same sampling procedure, we were able to reproduce their results; but we do not use this setting in our evaluation, as it is a simpler task than FewRel with its 80 different relations.

Finally, we notice a very small difference in performance between BERT and RoBERTa embeddings with \promptore{}. In practice both PLMs are well suited to provide precise results, and we decide to use BERT embeddings for the next parts of this paper.

\paragraph{Performance on domain specific datasets}
BERT and more broadly PLMs are usually pretrained on general domain data (e.g. Wikipedia), and we can ask ourselves if that impacts performances on "out of domain" datasets such as FewRel NYT and FewRel PubMed. We can see in Table \ref{tab:global_results} that \promptore{} does not see its results plummet. On the contrary, it still outperforms previous SOTA models by a large margin. \textbf{SelfORE}, which also relies on BERT embeddings, does not see its performance deteriorate as well, which seems to indicate BERT's ability to encode tokens not seen before.

In general, the results are higher than with FewRel, but that is explained by the fact that the two datasets contain less relations and instances.

Finally, we notice that \textbf{Etype+} predicts a single class on FewRel PubMed (as V-measure and ARI touch zero). As we have stated earlier, it is explained by the entity type schema, which is very limited as there are no \textit{Person}, \textit{Organization} or \textit{Location} entities in this dataset.

\begin{table}[tb]
    \centering
    \caption{Results of PromptORE with different prompts. PromptORE is trained with the exact number of relations $k$.}
    \label{tab:alternative_prompts}
    \small
    \begin{tabular}{r|l|ccc}
        \toprule
        Dataset & Prompt & \bcube{} (F1) & V (F1) & ARI \\
        \midrule
        \multirow{5}{*}{FewRel} & PromptORE ($\prompt$) & 48.8 & 71.8 & 43.4 \\
        & \qquad$\prompt_{\emptyset}$ & 33.8 & 57.4 & 28.8 \\
        & \qquad$\prompt'_{1}$ & 48.9 & 71.7 & 44.5 \\
        & \qquad$\prompt'_{2}$ & 49.4 & 72.4 & 46.3 \\
        & \qquad$\prompt'_{3}$ & 50.5 & 73.0 & 47.7 \\
        \midrule
        \multirow{5}{*}{\shortstack{FewRel \\ NYT}} & PromptORE ($\prompt$) & 65.1 & 78.0 & 56.9 \\
        & \qquad$\prompt_{\emptyset}$ & 51.3 & 65.7 & 41.6 \\
        & \qquad$\prompt'_{1}$ & 65.8 & 77.8 & 62.0 \\
        & \qquad$\prompt'_{2}$ & 61.0 & 74.8 & 56.9 \\
        & \qquad$\prompt'_{3}$ & 65.6 & 77.7 & 61.7 \\
        \midrule
        \multirow{5}{*}{\shortstack{FewRel \\ PubMed}} & PromptORE ($\prompt$) & 77.4 & 81.1 & 73.8 \\
        & \qquad$\prompt_{\emptyset}$ & 62.0 & 66.2 & 53.1 \\
        & \qquad$\prompt'_{1}$ & 76.4 & 80.0 & 72.3 \\
        & \qquad$\prompt'_{2}$ & 76.0 & 80.0 & 72.9 \\
        & \qquad$\prompt'_{3}$ & 77.4 & 81.1 & 73.1 \\
        \bottomrule
    \end{tabular}
\end{table}

\paragraph{Does PromptORE really extract relations?}
The core of \promptore{} is its prompt $\prompt$ that is used by the Relation Encoder to embed each instance. However, one can ask if BERT really uses the text of the current instance to predict the missing token (and thus extracts information from the sentence), or if it is only using its internal knowledge, ignoring the current instance context. To answer this question, we propose to create an empty prompt $\prompt_{\emptyset}$ where we do not input the current instance text. Its template is defined as:
\begin{equation}
    \prompt_{\emptyset}(\S, \eone, \etwo) = \texttt{"\cls{} $\eone$ \mask{} $\etwo.$\sep{}"}
\end{equation}
It is equivalent to $\prompt$ defined in eq. (\ref{eq:prompt}), except that we have removed $\S$.

The results are shown in Table \ref{tab:alternative_prompts}. We can see that the performance for all three metrics and three datasets are much lower with $\prompt_{\emptyset}$ compared to $\prompt$, with an average gap of 15\% in \bcube{}, 14\% in V-measure and 15\% in ARI. Therefore, it shows that BERT really benefits from the instance context to extract more precisely the relation between the two entities. It is interesting to remark that even without the instance text, \promptore{} still surpasses \textbf{SelfORE}, which clearly indicates that \textbf{SelfORE} fails to take full advantage of BERT embeddings.

\paragraph{Alternative Prompts} \label{sec:alternative_prompts_results}
As we have discussed in section \ref{sec:alternative_prompts}, $\prompt$ is not necessarily the best prompt, as more relations can be expressed with a noun than with a verb. We computed \promptore{} performances with three alternative prompts:
$\prompt'_{1}$, which encourages BERT to predict a noun,
$\prompt'_{2}$ with the prefix proposed by \cite{Lv2022}, and
$\prompt'_{3}$ containing a prefix variant of $\prompt'_{2}$.
Results are shown in Table \ref{tab:alternative_prompts}.

First, we notice that $\prompt'_{1}$ provides better results than $\prompt$ in ARI, but similar performances in V-measure and \bcube{} for FewRel and FewRel NYT. No improvement is observed with FewRel PubMed. This result is interesting because we showed that fewer relations can be expressed with a verb than with a noun, so we expected a gap in favor of $\prompt'_{1}$. For example FewRel relations \texttt{instance\_of}, \texttt{competition\_class}, \texttt{constellation} or \texttt{operating\_system} cannot be expressed with a verb but are nonetheless correctly identified with $\prompt$. It seems that BERT is weakly impacted by the apparent impossibility to predict a meaningful word.

In Table \ref{tab:alternative_prompts}, $\prompt'_{2}$ and $\prompt'_{3}$ achieve higher performances than $\prompt'_{1}$ in the majority of the cases, while their only difference with $\prompt'_{1}$ is the prefix (\textit{In this sentence} or \textit{We deduce that}). We can also see the impact of prompt's wording: at a first glance both prefixes seem to convey the same idea, but their performances are different. In fact if we replace \textit{deduce} by \textit{conclude} in $\prompt'_{3}$ we obtain lower performances (not shown in Table \ref{tab:alternative_prompts}). 

Finally, we notice that there is no consensus on the best prompt from the four proposed ones: $\prompt'_{3}$ is the best for FewRel, $\prompt'_{2}$ for FewRel NYT and $\prompt$ for FewRel PubMed. This highlights the importance to select and fine-tune prompts to maximize BERT's performances, which is indeed a major research area for prompt-based methods \cite{Lv2022, Shin2020, Haviv2021, Jiang2020}.  Under our fully unsupervised setting's goal, it is unfeasible to fine-tune the prompt due to the lack of labeled data, therefore we decide to keep our original $\prompt$ for the sake of fair results.

\subsection{Clustering without knowing \texorpdfstring{$k$}{k}}
Up to now, \promptore{} has access to $k$, the number of different relations. However, as said in section \ref{sec:without_k}, the most general setting is when we do not know $k$. We identified two methods to cluster our data without $k$: 
\begin{inparaenum}
    \item OPTICS, a clustering algorithm based on density, and
    \item the Elbow Rule (to compute $\hat{k}$ an estimation of $k$) with k-means clustering.
\end{inparaenum}
The results are shown in Table \ref{tab:without_k}.

\begin{table}[tb]
    \centering
    \caption{Results of PromptORE using different methods to estimate $k$. \textit{"Ideal"} represents results when $k$ is provided.}
    \label{tab:without_k}
    \small
    \begin{tabular}{r|l|c|ccc}
    \toprule
    Dataset & Method & $\hat{k}$ & \bcube{} (F1) & V (F1) & ARI \\
    \midrule
    \multirow{3}{*}{FewRel} & \textit{Ideal} & \textit{80} & 48.8 & \textbf{71.8} & \textbf{43.4} \\
    & OPTICS & 571 & 10.8 & 8.5 & 0 \\
    & Elbow & 65 & \textbf{49.5} & 71.2 & 42.2 \\
    \midrule
    \multirow{3}{*}{\shortstack{FewRel \\ NYT}} & \textit{Ideal} & \textit{25} & \textbf{65.1} & \textbf{78.0} & \textbf{56.9} \\
    & OPTICS & 35 & 33.2 & 29.3 & 1.7 \\
    & Elbow & 26 & 64.1 & 77.4 & 56.2 \\
    \midrule
    \multirow{3}{*}{\shortstack{FewRel \\ PubMed}} & \textit{Ideal} & \textit{10} & \textbf{77.4} & \textbf{81.1} & \textbf{73.8} \\
    & OPTICS & 12 & 26.8 & 11.2 & 0.3 \\
    & Elbow & 10 & \textbf{77.4} & \textbf{81.1} & \textbf{73.8}\\
    \bottomrule
    \end{tabular}
\end{table}

\begin{figure}[tbh]
    \centering
    \input{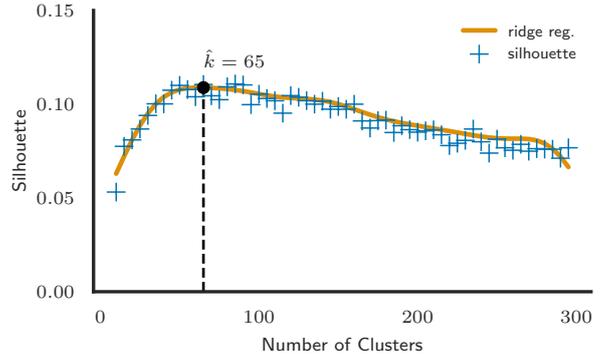}
    \caption{Results of the Elbow Rule on FewRel. Estimated number of relations $\hat{k}$ is equal to 65.}
    \label{fig:elbow}
\end{figure}

To give technical details, we use the \texttt{scikit-learn} implementation of OPTICS and k-means. To apply the Elbow Rule, we first calculate multiple clusterings by varying the number of clusters. For each of these clusterings we compute the silhouette coefficient. We obtain the blue scatter plot of the Figure \ref{fig:elbow} for FewRel. As this plot is rough, we approximate it thanks to a ridge regression with a gaussian kernel (orange curve in Figure \ref{fig:elbow}). In our case, this curve has a maximum, it is therefore easy to locate the elbow. We noticed the same curve shape with a maximum for FewRel NYT and FewRel PubMed. Sometimes however, it is possible to observe a growing curve, in which case automatic approaches \cite{Kolesnichenko2021} can be applied to locate the elbow. We detect the elbow at $\hat{k}=65$ clusters for FewRel. We obtain $\hat{k}=26$ for FewRel NYT and $\hat{k}=10$ for FewRel PubMed, that is, values of $\hat{k}$ nearly identical to the real number of relations.

\paragraph{Quantitative Results} OPTICS sets $\hat{k}$ at $571$ (see Table \ref{tab:without_k}), far from the optimal $k=80$ for FewRel. It translates into very poor performances compared to \promptore{} when we know $k$. We also note that OPTICS is very slow during training ($\sim$ 6h compared to 5min with k-means). On the other side, results are much more satisfactory with the Elbow Rule with a slight decrease in ARI but equivalent performances in \bcube{} and V-measure. Training time is also much more reasonable with $\sim$ 1h. We make the same conclusion when we look at FewRel NYT and FewRel PubMed.

We can conclude that, at least on our three datasets, the Elbow Rule is effective to find a correct estimation of $k$.

Finally, it is interesting to see that \promptore{} with the Elbow Rule widely surpasses previous SOTA approaches (Table \ref{tab:global_results}), with a gap of 15-25\% in \bcube{} and V-measure and 17-30\% in ARI. In our opinion, we demonstrate that it is possible to remove the dependency on all hyperparameters (including $k$) and still achieve state-of-the-art results.

\begin{figure}
    \centering
    \input{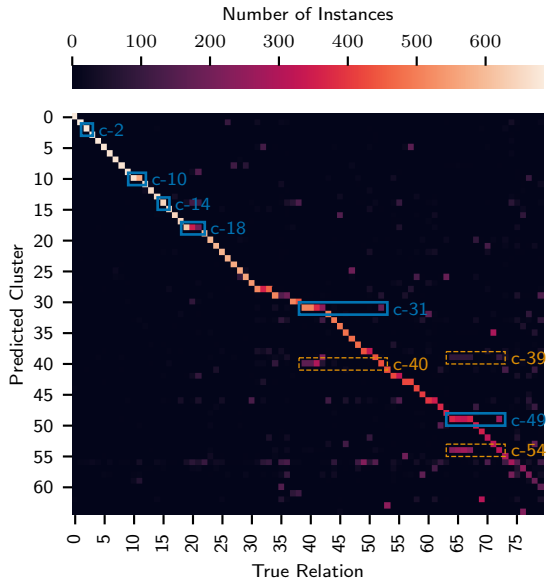}
    \caption{Confusion matrix between the true relations and the clusters found by PromptORE with the Elbow Rule on FewRel. The main relations of some clusters are highlighted.}
    \label{fig:elbow_cm}
\end{figure}

\paragraph{Qualitative Analysis of the Clustering} From Table \ref{tab:without_k}, we know that the Elbow Rule finds $\hat{k}=65$ instead of $80$ for FewRel. It means that the clustering cannot be ideal: some clusters must contain multiple relations. The confusion matrix between the true relations and the ones predicted by \promptore{} is shown in Figure \ref{fig:elbow_cm}. It is obviously not square as the number of clusters $\hat{k}$ is not equal to the number of relations $k$. We reorganize the axes to find a logical representation of the confusion matrix (as initially there is no link between the relation ids and the cluster ids).

On this confusion matrix, we notice indeed that some clusters are not pure: they contain multiple relations (e.g. clusters c-10, c-18, c-28, c-29, c-30, c-31, c-49, or c-54). The main observation is nevertheless that the matrix possesses a clear diagonal, meaning that \promptore{} is able to effectively distinguish the vast majority of the relations, while training in a fully unsupervised setting.

We also see that clusters seems to be relatively complete: there are seldom clusters sharing the same relations except for clusters c-31 and c-40; c-39, c-49 and c-54.

As some clusters contain multiple relations, we find it interesting to check whether the relations that compose each of these clusters are semantically linked. We randomly sample four clusters that contain multiple relations. Results are shown in Table \ref{tab:cluster_content}. For these four clusters, we can see that the relations are indeed semantically close within a cluster: 
\begin{itemize}
    \item for cluster c-10 they are linked to language,
    \item for cluster c-18 to geographical location,
    \item for cluster c-31 to artistic creation,
    \item for cluster c-49 to family relationship.
\end{itemize}
Even though these clusters are not optimal from the FewRel annotation point of view, they are semantically coherent. In fact, we could even argue that cluster c-10 makes more sense than the initial labeling which divided this cluster in two relations.

In conclusion, this qualitative analysis shows that the combination of \promptore{} with the Elbow Rule efficiently discovers semantically consistent clusters, which are very close to true relations.

\begin{table}[thb]
    \caption{Relation distributions that compose four randomly sampled impure clusters.}
    \label{tab:cluster_content}
    \centering
    \small
    \begin{tabularx}{0.45\textwidth}{c|X}
    \toprule
    Cluster & True Relations \\
    \midrule
    c-10 & language of film or TV show, \newline language of work or name \\
    \midrule
    c-18 & mountain range, \qquad located in physical feature, \newline located in or next to body of water \\
    \midrule
    c-31 & screenwriter, \qquad director, \qquad after a work by, \newline characters, \qquad composer \\
    \midrule
    c-49 & sibling, \qquad father, \qquad child, \qquad mother, \newline spouse \\
    \bottomrule
    \end{tabularx}
\end{table}

\subsection{Analysis of \texorpdfstring{$\prompt$}{P} Prompt Predictions}
In section \ref{sec:alternative_prompts_results}, we found a very little performance difference between $\prompt{}$ and $\prompt{}'_{1}$, while the majority of relations cannot be written using a verb. To go further, it would be interesting to check which tokens/verbs seem to describe best the clusters identified by \promptore{} with $\prompt$. To do that, we use the relation embeddings (computed by our Relation Encoder) and predict the masked tokens (represented by \mask{}) with the MLM head of BERT. By iterating for every instance located in one cluster, we can find the most frequent tokens that seem to describe it. We apply this method on the clusters identified by \promptore{} and Elbow Rule (see Figure \ref{fig:elbow_cm}). The results on three selected clusters are displayed in Table \ref{tab:cluster_naming}.

We observe that in most cases, the predicted names are not clear enough to qualify the relation corresponding to the cluster. Nevertheless, the names give clues to identify the general theme of the relation (\textit{married} indicates a family centered relation, \textit{borders} and \textit{surrounds} a geographic topic).

Table \ref{tab:cluster_naming} shows however the major limitation of $\prompt$: when we look at cluster c-49, \texttt{spouse} is represented by \textit{married}, but \texttt{sibling}, \texttt{father}, \texttt{child} and \texttt{mother} relations are not brought to light with the predicted names. Indeed, these four relations cannot be written using a single verb; by not finding satisfactory names, BERT defaulted into predicting punctuation tokens. We reach the same conclusion for cluster c-14, where BERT also predicted punctuation tokens.

\begin{table}[tb]
    \caption{Most frequent predicted tokens for three different clusters identified by PromptORE for FewRel. True relations composing these clusters are also displayed.}
    \label{tab:cluster_naming}
    \small
    \centering
    \begin{tabularx}{0.45\textwidth}{c|s|X}
        \toprule
        Cluster & Predicted tokens & True relations \\
        \midrule
        c-2 & , \newline for \newline supports & 99\%: platform \newline \textit{other}\\
        \midrule
        c-14 & : \newline borders \newline , \newline surrounds & 79\%: contains administrative territory \newline 5\%: located in administrative entity \newline 3\%: located in physical feature \newline \textit{other} \\
        \midrule
        c-49 & , \newline . \newline married & 25\%: sibling \newline 21\%: father \newline 19\%: child \newline 18\%: mother \newline 17\%: spouse\\
        \bottomrule
    \end{tabularx}
\end{table}

Finally, this observation gives us interesting insights on the behavior of our Relation Encoder. Intuitively, we could think that \promptore{} would have poor results with relations that cannot be written using a verb. On the contrary, we found that results were close between $\prompt$ and $\prompt'_{1}$ (Table \ref{tab:alternative_prompts}). At the same time, we notice that cluster c-2 (Table \ref{tab:cluster_naming}) is very pure, yet its most predicted name is \textit{","} (a token that is furthermore shared among the two other clusters of Table \ref{tab:cluster_naming}). In our opinion, it indicates that BERT is able to encode a very expressive embedding of the current relation instance that allows a precise clustering, but that cannot be translated into real words. This is supported by the fact that \promptore{} identified three different clusters with punctuation as their most frequent tokens (Table \ref{tab:cluster_naming}). It comforts us in the idea that complex prompts are not required to effectively represent a significant number of relations. It does not undermine however the importance of prompt-tuning: as showed in Table \ref{tab:alternative_prompts}, prompts have an impact on model performances.

\section{Conclusion and Future Work}
In this work, we introduce \promptore{}, an unsupervised RE model. Our proposed approach leverages and adapts the novel prompt-tuning paradigm. Experiments on one general and two domain specific datasets show that \promptore{} surpasses previous state-of-the-art methods, while being simpler and not needing hyperparameter tuning. On a secondary note, finding descriptive names for the clusters is still an open question.

In the future, we plan to explore other clustering approaches with a focus on Deep Clustering methods (e.g. \cite{Ronen2022}), and Hierarchical Clustering models that leverage the hierarchical nature of relations. We further envision to \textit{close the loop} of knowledge extraction, that is, benefiting from \promptore{}'s ability to extract relations in order to build a knowledge graph that can be used to further improve \promptore{}'s predictions.

\section*{Acknowledgments}
This work is supported by Alteca and the French Association for Research and Technology (ANRT) under CIFRE PhD fellowship n°2021/0851.

\bibliographystyle{plain}
\bibliography{references}

\begin{thebibliography}{10}

\bibitem{Ankerst1999}
Mihael Ankerst, Markus~M. Breunig, Hans~Peter Kriegel, and Jörg Sander.
\newblock {OPTICS: Ordering Points to Identify the Clustering Structure}.
\newblock {\em SIGMOD Record (ACM Special Interest Group on Management of
  Data)}, (2):49--60, 6 1999.

\bibitem{Baldwin1998}
Amit Bagga and Breck Baldwin.
\newblock {Algorithms for scoring coreference chain}.
\newblock In {\em Proceedings of the 1st International Conf. on Language
  Resources and Evaluation Workshop on Linguistics Coreference}, page
  563–566, 1998.

\bibitem{Banko2007}
Michele Banko, Michael~J Cafarella, Stephen Soderland, Matt Broadhead, and Oren
  Etzioni.
\newblock {Open Information Extraction from the Web}.
\newblock In {\em Proceedings of the 20th International Joint Conf. on
  Artificial Intelligence}, pages 2670--2676, Hyderabad, India, 2007. Morgan
  Kaufmann Publishers Inc.

\bibitem{Blei2003}
David~M Blei, Andrew~Y Ng, and Michael~I. Jordan.
\newblock {Latent Dirichlet allocation}.
\newblock {\em Journal of Machine Learning Research}, (4-5):993--1022, 2003.

\bibitem{Chen2022}
Xiang Chen, Ningyu Zhang, Xin Xie, Shumin Deng, Yunzhi Yao, Chuanqi Tan, Fei
  Huang, Luo Si, and Huajun Chen.
\newblock {KnowPrompt: Knowledge-aware Prompt-tuning with Synergistic
  Optimization for Relation Extraction}.
\newblock In {\em Proceedings of the ACM Web Conf. 2022}, Lyon, France, 4 2022.
  ACM.

\bibitem{Chen2020}
Xiaojun Chen, Shengbin Jia, and Yang Xiang.
\newblock {A review: Knowledge reasoning over knowledge graph}.
\newblock {\em Expert Systems with Applications}, page 112948, 3 2020.

\bibitem{Chopra2005}
Sumit Chopra, Raia Hadsell, and Yann LeCun.
\newblock {Learning a similarity metric discriminatively, with application to
  face verification}.
\newblock In {\em Proceedings of the 2005 IEEE Computer Society Conf. on
  Computer Vision and Pattern Recognition}, pages 539--546, San Diego, CA, USA,
  2005. IEEE Computer Society.

\bibitem{Cui2018}
Lei Cui, Furu Wei, and Ming Zhou.
\newblock {Neural open information extraction}.
\newblock In {\em Proceedings of the 56th Annual Meeting of the ACL}, pages
  407--413, Melbourne, Australia, 2018. ACL.

\bibitem{Devlin2019}
Jacob Devlin, Ming~Wei Chang, Kenton Lee, and Kristina Toutanova.
\newblock {BERT: Pre-training of deep bidirectional transformers for language
  understanding}.
\newblock In {\em Proceedings of the 2019 Conf. of the NAACL: Human Language
  Technologies}, pages 4171--4186, Stroudsburg, PA, USA, 2019. ACL.

\bibitem{Elsahar2017}
Hady Elsahar, Elena Demidova, Simon Gottschalk, Christophe Gravier, and
  Frederique Laforest.
\newblock {Unsupervised Open Relation Extraction}.
\newblock In {\em The Semantic Web: ESWC 2017 Satellite Events}, pages 12--16,
  Cham, 1 2017. Springer.

\bibitem{Ester1996}
Martin Ester, Hans-Peter Kriegel, Jörg Sander, and Xiaowei Xu.
\newblock {A Density-Based Algorithm for Discovering Clusters in Large Spatial
  Databases with Noise}.
\newblock In {\em Proceedings of the 2nd International Conf. on Knowledge
  Discovery and Data Mining}, pages 226--231, Portland, Oregon, United States,
  1996. AAAI Press.

\bibitem{Frey2007}
Brendan~J. Frey and Delbert Dueck.
\newblock {Clustering by passing messages between data points}.
\newblock {\em Science}, (5814):972--976, 2 2007.

\bibitem{Gao2021}
Tianyu Gao, Adam Fisch, and Danqi Chen.
\newblock {Making pre-trained language models better few-shot learners}.
\newblock In {\em Proceedings of the 59th Annual Meeting of the ACL and the
  11th International Joint Conf. on Natural Language Processing}, pages
  3816--3830, Online, 12 2021. ACL.

\bibitem{Gao2019}
Tianyu Gao, Xu~Han, Hao Zhu, Zhiyuan Liu, Peng Li, Maosong Sun, and Jie Zhou.
\newblock {Fewrel 2.0: Towards more challenging few-shot relation
  classification}.
\newblock In {\em Proceedings of the 2019 Conf. on Empirical Methods in Natural
  Language Processing and 9th International Joint Conf. on Natural Language
  Processing}, pages 6250--6255, Hong Kong, China, 2019. ACL.

\bibitem{Gong2021a}
Jiaying Gong and Hoda Eldardiry.
\newblock {Prompt-based Zero-shot Relation Classification with Semantic
  Knowledge Augmentation}, 12 2021.

\bibitem{Guo2020}
Qingyu Guo, Fuzhen Zhuang, Chuan Qin, Hengshu Zhu, Xing Xie, Hui Xiong, and
  Qing He.
\newblock {A Survey on Knowledge Graph-Based Recommender Systems}.
\newblock {\em IEEE Transactions on Knowledge and Data Engineering}, pages
  1--17, 10 2020.

\bibitem{Han2020}
Xu~Han, Tianyu Gao, Yankai Lin, Hao Peng, Yaoliang Yang, Chaojun Xiao, Zhiyuan
  Liu, Peng Li, Maosong Sun, and Jie Zhou.
\newblock {More Data, More Relations, More Context and More Openness: A Review
  and Outlook for Relation Extraction}.
\newblock In {\em Proceedings of the 1st Conf. of the Asia-Pacific Chapter of
  the ACL and the 10th International Joint Conf. on Natural Language
  Processing}, pages 745--758, Suzhou, China, 2020. ACL.

\bibitem{Han2019}
Xu~Han, Tianyu Gao, Yuan Yao, Demin Ye, Zhiyuan Liu, and Maosong Sun.
\newblock {OpenNRE: An open and extensible toolkit for neural relation
  extraction}.
\newblock In {\em Proceedings of the 2019 Conf. on Empirical Methods in Natural
  Language Processing and the 9th International Joint Conf. on Natural Language
  Processing: System Demonstrations}, pages 169--174, Hong Kong, China, 2019.
  ACL.

\bibitem{Han2018}
Xu~Han, Hao Zhu, Pengfei Yu, Ziyun Wang, Yuan Yao, Zhiyuan Liu, and Maosong
  Sun.
\newblock {Fewrel: A large-scale supervised few-shot relation classification
  dataset with state-of-the-art evaluation}.
\newblock In {\em Proceedings of the 2018 Conf. on Empirical Methods in Natural
  Language Processing}, pages 4803--4809, Brussels, Belgium, 2018. ACL.

\bibitem{Haviv2021}
Adi Haviv, Jonathan Berant, and Amir Globerson.
\newblock {BERTese: Learning to speak to BERT}.
\newblock In {\em Proceedings of the 16th Conf. of the European Chapter of the
  ACL}, pages 3618--3623, Online, 2021. ACL.

\bibitem{Hu2020}
Xuming Hu, Lijie Wen, Yusong Xu, Chenwei Zhang, and Philip~S. Yu.
\newblock {SelfORE: Self-supervised relational feature learning for open
  relation extraction}.
\newblock In {\em Proceedings of the 2020 Conf. on Empirical Methods in Natural
  Language Processing}, pages 3673--3682, Online, 2020. ACL.

\bibitem{Huang2019}
Xiao Huang, Jingyuan Zhang, Dingcheng Li, and Ping Li.
\newblock {Knowledge graph embedding based question answering}.
\newblock In {\em Proceedings of the 12th ACM International Conf. on Web Search
  and Data Mining}, pages 105--113, Melbourne, Australia, 1 2019. ACM, Inc.

\bibitem{Hubert1985}
Lawrence Hubert and Phipps Arabie.
\newblock {Comparing partitions}.
\newblock {\em Journal of Classification}, (1):193--218, 12 1985.

\bibitem{Ji2021}
Shaoxiong Ji, Shirui Pan, Erik Cambria, Senior Member, Pekka Marttinen,
  Philip~S. Yu, and Life Fellow.
\newblock {A Survey on Knowledge Graphs: Representation, Acquisition, and
  Applications}.
\newblock {\em IEEE Transactions on Neural Networks and Learning Systems},
  pages 1 -- 21, 2021.

\bibitem{Jiang2020}
Zhengbao Jiang, Frank~F. Xu, Jun Araki, and Graham Neubig.
\newblock {How can we know what language models know?}
\newblock {\em Transactions of the ACL}, pages 423--438, 2020.

\bibitem{Kingma2014}
Diederik~P. Kingma and Max Welling.
\newblock {Auto-encoding variational bayes}.
\newblock In {\em Proceedings of the 2nd International Conf. on Learning
  Representations}, Banff, Canada, 12 2014. International Conf. on Learning
  Representations.

\bibitem{Kolesnichenko2021}
Pavel~V Kolesnichenko, Qianhui Zhang, Changxi Zheng, Michael~S Fuhrer, and
  Jeffrey~A Davis.
\newblock {Multidimensional analysis of excitonic spectra of monolayers of
  tungsten disulphide: toward computer-aided identification of structural and
  environmental perturbations of 2D materials}.
\newblock {\em Machine Learning: Science and Technology}, (2):025021, 3 2021.

\bibitem{Lin2020}
Ying Lin, Heng Ji, Fei Huang, and Lingfei Wu.
\newblock {A Joint Neural Model for Information Extraction with Global
  Features}.
\newblock In {\em Proceedings of the 58th Annual Meeting of the ACL}, pages
  7999--8009, Online, 7 2020. ACL.

\bibitem{Liu2021}
Fangchao Liu, Lingyong Yan, Hongyu Lin, Xianpei Han, and Le~Sun.
\newblock {Element intervention for open relation extraction}.
\newblock In {\em Proceedings of the 59th Annual Meeting of the ACL and the
  11th International Joint Conf. on Natural Language Processing}, pages
  4683--4693, Online, 6 2021. ACL.

\bibitem{Liu2021a}
Pengfei Liu, Weizhe Yuan, Jinlan Fu, Zhengbao Jiang, Hiroaki Hayashi, and
  Graham Neubig.
\newblock {Pre-train, Prompt, and Predict: A Systematic Survey of Prompting
  Methods in Natural Language Processing}, 7 2021.

\bibitem{Liu2019}
Yinhan Liu, Myle Ott, Naman Goyal, Jingfei Du, Mandar Joshi, Danqi Chen, Omer
  Levy, Mike Lewis, Luke Zettlemoyer, and Veselin Stoyanov.
\newblock {RoBERTa: A Robustly Optimized BERT Pretraining Approach}, 7 2019.

\bibitem{Lloyd1957}
S.~P. Lloyd.
\newblock {Least squares quantization in PCM}.
\newblock {\em Technical Report RR-5497}, 1957.

\bibitem{Lou2021}
Renze Lou, Fan Zhang, Xiaowei Zhou, Yutong Wang, Minghui Wu, and Lin Sun.
\newblock {A Unified Representation Learning Strategy for Open Relation
  Extraction with Ranked List Loss}.
\newblock In {\em Proceedings of the 20th China National Conf. on Computational
  Linguistics}, pages 1096--1108, Huhhot, China, 2021. Chinese Information
  Processing Society of China.

\bibitem{Luan2019}
Yi~Luan, Dave Wadden, Luheng He, Amy Shah, Mari Ostendorf, and Hannaneh
  Hajishirzi.
\newblock {A general framework for information extraction using dynamic span
  graphs}.
\newblock In {\em Proceedings of the 2019 Conf. of the NAACL: Human Language
  Technologies}, pages 3036--3046, Minneapolis, Minnesota, 2019. ACL.

\bibitem{Lv2022}
Bo~Lv, Li~Jin, Yanan Zhang, Hao Wang, Xiaoyu Li, and Zhi Guo.
\newblock {Commonsense Knowledge-Aware Prompt Tuning for Few-Shot NOTA Relation
  Classification}.
\newblock {\em Applied Sciences}, (4):2185, 2 2022.

\bibitem{Observations1967}
James MacQueen.
\newblock {Some methods for classification and analysis of multivariate
  observations}.
\newblock In {\em Proceedings of the 5th Berkeley Symp. on mathematical
  statistics and probability}, pages 281--297, Berkeley, California, United
  States, 1967. University of California Press.

\bibitem{Marcheggiani2016}
Diego Marcheggiani and Ivan Titov.
\newblock {Discrete-State Variational Autoencoders for Joint Discovery and
  Factorization of Relations}.
\newblock {\em Transactions of the ACL}, pages 231--244, 12 2016.

\bibitem{Mintz2009}
Mike Mintz, Steven Bills, Rion Snow, and Dan Jurafsky.
\newblock {Distant supervision for relation extraction without labeled data}.
\newblock In {\em Proceedings of the Joint Conf. of the 47th Annual Meeting of
  the ACL and the 4th International Joint Conf. on Natural Language}, pages
  1003--1011, Suntec, Singapore, 2009. ACL.

\bibitem{Pennington2014}
Jeffrey Pennington, Richard Socher, and Christopher~D. Manning.
\newblock {GloVe: Global vectors for word representation}.
\newblock In {\em Proceedings of the 2014 Conf. on Empirical Methods in Natural
  Language Processing}, pages 1532--1543, Doha, Qatar, 2014. ACL.

\bibitem{Perez2021}
Ethan Perez, Douwe Kiela, and Kyunghyun Cho.
\newblock {True Few-Shot Learning with Language Models}, 5 2021.

\bibitem{Peters2018}
Matthew~E. Peters, Mark Neumann, Mohit Iyyer, Matt Gardner, Christopher Clark,
  Kenton Lee, and Luke Zettlemoyer.
\newblock {Deep contextualized word representations}.
\newblock In {\em Proceedings of the 2018 Conf. of the NAACL: Human Language
  Technologies}, pages 2227--2237, New Orleans, Louisiana, United States, 2
  2018. ACL.

\bibitem{Ren2020}
Haopeng Ren, Yi~Cai, Xiaofeng Chen, Guohua Wang, and Qing Li.
\newblock {A Two-phase Prototypical Network Model for Incremental Few-shot
  Relation Classification}.
\newblock In {\em Proceedings of the 28th International Conf. on Computational
  Linguistics}, pages 1618--1629, Barcelona, Spain, 1 2020. ACL.

\bibitem{Riedel2010}
Sebastian Riedel, Limin Yao, and Andrew McCallum.
\newblock {Modeling relations and their mentions without labeled text}.
\newblock In {\em Lecture Notes in Computer Science (including subseries
  Lecture Notes in Artificial Intelligence and Lecture Notes in
  Bioinformatics)}, number PART 3, pages 148--163, Berlin, Heidelberg, 2010.
  Springer.

\bibitem{Ronen2022}
Meitar Ronen, Shahaf~E. Finder, and Oren Freifeld.
\newblock {DeepDPM: Deep Clustering With an Unknown Number of Clusters}, 3
  2022.

\bibitem{Rosenberg2007}
Andrew Rosenberg and Julia Hirschberg.
\newblock {V-Measure: A conditional entropy-based external cluster evaluation
  measure}.
\newblock In {\em Proceedings of the 2007 Joint Conf. on Empirical Methods in
  Natural Language Processing and Computational Natural Language Learning},
  pages 410--420, Prague, Czech Republic, 2007. ACL.

\bibitem{Rousseeuw1987}
Peter~J. Rousseeuw.
\newblock {Silhouettes: A graphical aid to the interpretation and validation of
  cluster analysis}.
\newblock {\em Journal of Computational and Applied Mathematics}, (C):53--65,
  11 1987.

\bibitem{Roy2019}
Arpita Roy, Youngja Park, Taesung Lee, and Shimei Pan.
\newblock {Supervising unsupervised open information extraction models}.
\newblock In {\em Proceedings of the 2019 Conf. on Empirical Methods in Natural
  Language Processing and 9th International Joint Conf. on Natural Language
  Processing}, pages 728--737, Hong Kong, China, 2019. ACL.

\bibitem{Saha2018}
Swarnadeep Saha and {Mausam}.
\newblock {Open information extraction from conjunctive sentences}.
\newblock In {\em Proceedings of the 27th International Conf. on Computational
  Linguistics}, pages 2288--2299, Santa Fe, New Mexico, USA, 2018. ACL.

\bibitem{Shin2020}
Taylor Shin, Yasaman Razeghi, Robert~L. Logan, Eric Wallace, and Sameer Singh.
\newblock {AUTOPROMPT: Eliciting knowledge from language models with
  automatically generated prompts}.
\newblock In {\em Proceedings of the 2020 Conf. on Empirical Methods in Natural
  Language Processing}, pages 4222--4235, Online, 10 2020. ACL.

\bibitem{Sibson1973}
R.~Sibson.
\newblock {SLINK: An optimally efficient algorithm for the single-link cluster
  method}.
\newblock {\em The Computer Journal}, (1):30--34, 1 1973.

\bibitem{Simon2019}
Étienne Simon, Vincent Guigue, and Benjamin Piwowarski.
\newblock {Unsupervised information extraction: Regularizing discriminative
  approaches with relation distribution losses}.
\newblock In {\em Proceedings of the 57th Annual Meeting of the ACL}, pages
  1378--1387, Florence, Italy, 2019. ACL.

\bibitem{Snell2017}
Jake Snell, Kevin Swersky, and Richard Zemel.
\newblock {Prototypical networks for few-shot learning}.
\newblock In {\em Advances in Neural Information Processing Systems}, pages
  4078--4088. Neural information processing systems foundation, 3 2017.

\bibitem{Soares2019}
Livio~Baldini Soares, Nicholas FitzGerald, Jeffrey Ling, and Tom Kwiatkowski.
\newblock {Matching the blanks: Distributional similarity for relation
  learning}.
\newblock In {\em Proceedings of the 57th Annual Meeting of the ACL}, pages
  2895--2905, Florence, Italy, 2019. ACL.

\bibitem{Stanovsky2018}
Gabriel Stanovsky, Julian Michael, Luke Zettlemoyer, and Ido Dagan.
\newblock {Supervised open information extraction}.
\newblock In {\em Proceedings of the 2018 Conf. of the NAACL: Human Language
  Technologies}, pages 885--895, New Orleans, Louisiana, United States, 2018.
  ACL.

\bibitem{Steinley2004}
Douglas Steinley.
\newblock {Properties of the Hubert-Arabie adjusted Rand index}.
\newblock {\em Psychological Methods}, (3):386--396, 9 2004.

\bibitem{Tan2022}
Jiejun Tan, Wenbin Hu, and WeiWei Liu.
\newblock {EPPAC: Entity Pre-typing Relation Classification with Prompt Answer
  Centralizing}, 3 2022.

\bibitem{Thorndike1953}
Robert~L. Thorndike.
\newblock {Who belongs in the family?}
\newblock {\em Psychometrika}, (4):267--276, 12 1953.

\bibitem{Tran2020b}
Thy~Thy Tran, Phong Le, and Sophia Ananiadou.
\newblock {Revisiting Unsupervised Relation Extraction}.
\newblock In {\em Proceedings of the 58th Annual Meeting of the ACL}, pages
  7498--7505, Online, 7 2020. ACL.

\bibitem{Vaswani2017}
Ashish Vaswani, Noam Shazeer, Niki Parmar, Jakob Uszkoreit, Llion Jones,
  Aidan~N. Gomez, Łukasz Kaiser, and Illia Polosukhin.
\newblock {Attention is all you need}.
\newblock In {\em Proceedings of Advances in Neural Information Processing
  Systems 30}, pages 5999--6009. Neural information processing systems
  foundation, 6 2017.

\bibitem{Wadden2019}
David Wadden, Ulme Wennberg, Yi~Luan, and Hannaneh Hajishirzi.
\newblock {Entity, relation, and event extraction with contextualized span
  representations}.
\newblock In {\em Proceedings of the 2019 Conf. on Empirical Methods in Natural
  Language Processing and 9th International Joint Conf. on Natural Language
  Processing}, pages 5784--5789, Hong Kong, China, 2019. ACL.

\bibitem{Wang2019a}
Xinshao Wang, Yang Hua, Elyor Kodirov, Guosheng Hu, Romain Garnier, and Neil~M.
  Robertson.
\newblock {Ranked list loss for deep metric learning}.
\newblock In {\em Proceedings of the 2019 IEEE Computer Society Conf. on
  Computer Vision and Pattern Recognition}, pages 5202--5211, Long Beach, CA,
  United States, 3 2019. IEEE Computer Society.

\bibitem{Wu2019}
Ruidong Wu, Yuan Yao, Xu~Han, Ruobing Xie, Zhiyuan Liu, Fen Lin, Leyu Lin, and
  Maosong Sun.
\newblock {Open relation extraction: Relational knowledge transfer from
  supervised data to unsupervised data}.
\newblock In {\em Proceedings of the 2019 Conf. on Empirical Methods in Natural
  Language Processing and 9th International Joint Conf. on Natural Language
  Processing}, pages 219--228, Hong Kong, China, 2019. ACL.

\bibitem{Yao2011}
Limin Yao, Aria Haghighi, Sebastian Riedel, and Andrew McCallum.
\newblock {Structured relation discovery using generative models}.
\newblock In {\em Proceedings of the 2011 Conf. on Empirical Methods in Natural
  Language Processing}, pages 1456--1466, Edinburgh, Scotland, UK, 2011. ACL.

\bibitem{Yao2012}
Limin Yao, Sebastian Riedel, and Andrew McCallum.
\newblock {Unsupervised relation discovery with sense disambiguation}.
\newblock In {\em Proceedings of the 50th Annual Meeting of the ACL}, pages
  712--720, Jeju Island, Korea, 2012. ACL.

\bibitem{Yuan2021}
Chenhan Yuan and Hoda Eldardiry.
\newblock {Unsupervised Relation Extraction: A Variational Autoencoder
  Approach}.
\newblock In {\em Proceedings of the 2021 Conf. on Empirical Methods in Natural
  Language Processing}, pages 1929--1938, Stroudsburg, PA, USA, 2021. ACL.

\bibitem{Zeng2015}
Daojian Zeng, Kang Liu, Yubo Chen, and Jun Zhao.
\newblock {Distant supervision for relation extraction via Piecewise
  Convolutional Neural Networks}.
\newblock In {\em Proceedings of the 2015 Conf. on Empirical Methods in Natural
  Language Processing}, pages 1753--1762, Lisbon, Portugal, 2015. ACL.

\bibitem{Zhang2021b}
Kai Zhang, Yuan Yao, Ruobing Xie, Xu~Han, Zhiyuan Liu, Fen Lin, Leyu Lin, and
  Maosong Sun.
\newblock {Open Hierarchical Relation Extraction}.
\newblock In {\em Proceedings of the 2021 Conf. of the NAACL: Human Language
  Technologies}, pages 5682--5693, Online, 6 2021. ACL.

\bibitem{Zhang2019}
Ningyu Zhang, Shumin Deng, Zhanlin Sun, Guanying Wang, Xi~Chen, Wei Zhang, and
  Huajun Chen.
\newblock {Long-tail relation extraction via knowledge graph embeddings and
  graph convolution networks}.
\newblock In {\em Proceedings of the 2019 Conf. of the NAACL: Human Language
  Technologies}, pages 3016--3025, Minneapolis, Minnesota, USA, 2019. ACL.

\bibitem{Zhang2021c}
Wenkai Zhang, Hongyu Lin, Xianpei Han, and Le~Sun.
\newblock {De-biasing Distantly Supervised Named Entity Recognition via Causal
  Intervention}.
\newblock In {\em Proceedings of the 59th Annual Meeting of the ACL and the
  11th International Joint Conf. on Natural Language Processing}, pages
  4803--4813, Online, 6 2021. ACL.

\bibitem{Zhang2021a}
Wenkai Zhang, Hongyu Lin, Xianpei Han, Le~Sun, Huidan Liu, Zhicheng Wei, and
  Nicholas~Jing Yuan.
\newblock {Denoising Distantly Supervised Named Entity Recognition via a
  Hypergeometric Probabilistic Model}.
\newblock {\em Proceedings of the 35th AAAI Conf. on Artificial Intelligence},
  (16):14481--14488, 6 2021.

\bibitem{Zhao2021}
Jun Zhao, Tao Gui, Qi~Zhang, and Yaqian Zhou.
\newblock {A Relation-Oriented Clustering Method for Open Relation Extraction}.
\newblock In {\em Proceedings of the 2021 Conf. on Empirical Methods in Natural
  Language Processing}, pages 9707--9718, Punta Cana, Dominican Republic, 2021.
  ACL.

\bibitem{Zheng2019}
Shun Zheng, Xu~Han, Yankai Lin, Peilin Yu, Lu~Chen, Ling Huang, Zhiyuan Liu,
  and Wei Xu.
\newblock {DIAG-NRE: A neural pattern diagnosis framework for distantly
  supervised neural relation extraction}.
\newblock In {\em Proceedings of the 57th Annual Meeting of the ACL}, pages
  1419--1429, Florence, Italy, 2019. ACL.

\bibitem{Zhong2021}
Zexuan Zhong and Danqi Chen.
\newblock {A Frustratingly Easy Approach for Entity and Relation Extraction}.
\newblock In {\em Proceedings of the 2021 Annual Conf. of the NAACL}, pages
  50--61, Online, 2021. ACL.

\end{thebibliography}
\end{document}